\documentclass[12pt]{article}

\usepackage{amsmath}
\usepackage{amssymb}
\usepackage{booktabs}
\usepackage{array}
\usepackage{multirow}
\usepackage{xcolor}
\usepackage{tikz}
\usepackage{placeins}
\usetikzlibrary{arrows.meta,backgrounds,fit,positioning,shapes.geometric}
\usepackage[a4paper,top=2.5cm,bottom=2.5cm,left=2.5cm,right=2.5cm]{geometry}
\usepackage[numbers,sort&compress]{natbib}
\usepackage{hyperref}
\usepackage{orcidlink}

\newcolumntype{L}[1]{>{\raggedright\arraybackslash}p{#1}}
\hypersetup{
    colorlinks=true,           
    citecolor=blue,            
    linkcolor=red,             
    urlcolor=blue,             
    bookmarksnumbered=true,    
    pdfpagemode=UseOutlines    
}

\newcommand{\datasetname}{FllumaOne}
\newcommand{\releasename}{FllumaOne-100K}
\newcommand{\datasetsize}{100,000}
\newcommand{\flluma}{Flluma}
\newcommand{\occ}{OpenCASCADE}

\title{\datasetname: A Code-Native Multimodal CAD Dataset with Executable Programs and Kernel-Validated Feature Histories}

\author{Jizong Zhan~\orcidlink{0009-0006-3762-0328}\thanks{\texttt{jizongzh@gmail.com}}}

\date{June 2026}

\begin{document}

\maketitle

\begin{abstract}
  Parametric computer-aided design records both final geometry and the ordered construction history that determines how a part can be edited. Datasets for editable CAD research should therefore expose modeling operations, parameters, and feature dependencies together with validated geometry. We introduce \datasetname, a code-native multimodal CAD dataset whose models are generated by executable Python programs in \flluma, a Qt/C++ \occ-based CAD system. Each sample aligns its program with a structured feature tree, a training-oriented intermediate representation, STEP geometry, a surface point cloud, natural-language descriptions, metadata, and eight canonical visible-edge renderings.

  The primary release, \releasename, contains \datasetsize{} accepted samples across four template-level complexity regimes. Programs are executed and retained only after kernel geometry, solid validity, and export checks; release reports also record modality completeness and split-level duplicate tests. A Qwen2.5-Coder-1.5B LoRA baseline trained on 80,000 samples achieves 99.98\% Python syntax validity, 99.97\% \flluma{} build success, and 99.14\% STEP-export validity on the held-out 10,000-sample test split. For the 9,909 predictions converted to surface point clouds, the mean normalized Chamfer Distance is 0.002124. The dataset supports conditioned CAD reconstruction, executable program synthesis, feature-tree prediction, B-Rep analysis, retrieval, design completion, and editable reverse engineering.
\end{abstract}

\medskip

\section{Introduction}

An editable CAD model is not defined by its final shape alone. Its engineering value also lies in the construction history: the sketches, features, parameters, and dependencies that determine how the part can be modified. A mesh or final B-Rep may reproduce the boundary accurately while omitting this information. That omission limits design reuse, variant creation, manufacturing-feature analysis, and reverse engineering, where the required output is a model that can be edited rather than only viewed~\citep{zhang2026large}.

Large geometry repositories have supported substantial progress in shape learning and B-Rep processing~\citep{chang2015shapenet,koch2019abc}. Construction-aware datasets subsequently introduced command sequences, sketch graphs, and human-authored modeling histories~\citep{seff2020sketchgraphs,willis2021fusion360gallery,wu2021deepcad}. More recent datasets connect these representations to language, images, point samples, native CAD files, or executable scripts~\citep{lv2025cadinstruct,dong2026histcad,pyatov2026cadfs}. The remaining practical challenge is to combine an inspectable construction record, replayable code, kernel-validated geometry, and aligned learning inputs without requiring a commercial CAD license.

We introduce \datasetname, a code-native multimodal CAD dataset for learning editable construction structure. Each sample has a canonical feature tree and a paired Python program executed by \flluma, a Qt/C++ CAD system built on \occ. The release also provides a compact training IR and derived geometric, visual, and textual observations. Keeping the construction record and replayable program together permits failures to be examined at the code, feature-history, kernel, and export levels.

The 100,000 samples are generated from a controlled library of parametric construction procedures and accepted only after geometric validation and export checks. A separate local LLM stage produces captions, comprehensive descriptions, and prompt variants from kernel-derived records without selecting parameters or altering geometry, keeping geometric generation deterministic.

The main contributions are:
\begin{itemize}
  \item \textbf{Paired construction representations.} A canonical feature tree, executable Python program, and compact training IR describe the same editable modeling history at complementary levels of abstraction.
  \item \textbf{Kernel-grounded multimodal release.} The construction record is aligned with STEP geometry, surface point clouds, canonical views, text, and validation metadata for 100,000 accepted samples.
  \item \textbf{Auditable generation and integrity checks.} The release reports geometry and export validation, split membership, modality completeness, and exact-duplicate checks over programs, construction records, IR, and solid signatures.
  \item \textbf{Executable baseline and evaluation tools.} A reproducible text-to-program baseline is evaluated from Python syntax through \flluma{} execution, solid validation, STEP export, and normalized surface Chamfer Distance.
\end{itemize}

\section{Related Work}
\label{sec:related}

\subsection{From Shape Collections to CAD Boundary Representations}

ModelNet, ShapeNet, and Thingi10K established large-scale learning from voxels, meshes, and object collections~\citep{wu2015modelnet,chang2015shapenet,zhou2016thingi10k}. Their strength is geometric and semantic coverage rather than editable modeling structure. Asset-oriented resources such as Objaverse++ and 3DCOMPAT++ extend visual, material, and multimodal annotations, but retain the final object as the principal representation~\citep{lin2025objaversepp,slim2025compatpp}.

CAD research requires more exact geometry. ABC made analytic B-Rep curves and surfaces available at million-model scale, enabling patch, topology, and differential-geometry learning~\citep{koch2019abc}. UV-Net demonstrated the value of operating directly on B-Rep faces and edges, while SolidGen, BrepGen, and BrepDiff treat boundary entities as generative targets~\citep{jayaraman2021uvnet,jayaraman2023solidgen,xu2024brepgen,lee2025brepdiff}. Operation annotations add partial construction semantics: CC3D-Ops assigns operation and step labels to B-Rep faces, and eCAD-Net recovers editable sketch--extrude sequences from final B-Reps~\citep{dupont2022cc3dops,zhang2025ecadnet}. These resources establish strong supervision for final geometry and topology; construction history remains a separate representation problem.

\subsection{Construction Histories and Constraints}

Fusion 360 Gallery introduced human-authored construction sequences and an execution environment for sequential reconstruction~\citep{willis2021fusion360gallery}. DeepCAD showed that substantially larger command-sequence collections can support generative modeling, albeit within a compact sketch--extrude language~\citep{wu2021deepcad}. SketchGraphs approaches design intent from the sketch side, representing geometric constraints as graphs over millions of 2D sketches~\citep{seff2020sketchgraphs}. These datasets established three complementary ingredients of editable CAD data: authentic histories, scalable sequence representations, and explicit geometric relations.

Conditioned CAD generation has since expanded the inputs and operations under study. Text2CAD aligns natural-language prompts with construction sequences, Drawing2CAD predicts sequences from vector drawings, and RLCAD studies reinforcement learning for command generation~\citep{khan2024text2cad,qin2025drawing2cad,yin2026rlcad}. HistCAD combines executable histories with explicit sketch constraints and native CAD compatibility, placing parameter editability and constraint preservation at the center of evaluation~\citep{dong2026histcad}. This line of work makes clear that single-pass geometric validity and editability are related but distinct properties.

\subsection{Executable CAD Programs}

Executable scripts provide a representation that code models can generate and engineers can inspect. CAD-Recode maps point clouds to CadQuery programs and uses large procedurally generated training data~\citep{rukhovich2025cadrecode}. Zero-to-CAD embeds code generation in an execution-and-repair loop and broadens the operation vocabulary beyond sketch and extrusion~\citep{ataei2026zerotocad}. CAD-Coder and FutureCAD further combine language-model generation with geometric feedback or B-Rep grounding~\citep{guan2025cadcoder,li2026futurecad}. The principal trade-off is that scale and executability are obtained through a chosen scripting API and generation procedure, whose conventions become part of the learned distribution.

Other work conditions construction on images, text, or cross-modal retrieval. GenCAD and DiffCAD study image-conditioned generation or alignment, while CAD-GPT and CADFusion integrate language, spatial reasoning, and visual feedback~\citep{alam2025gencad,gao2024diffcad,wang2025cadgpt,wang2025cadfusion}. Executable program data are therefore increasingly evaluated not only as code corpora but as a bridge between observations and editable geometry.

\subsection{Multimodal CAD Datasets}

Recent datasets package construction data with several observation modalities. CADInstruct adds language supervision to CAD programs, and Omni-CAD aligns command sequences with text, images, and point data~\citep{lv2025cadinstruct,xu2025cadmllm}. HistCAD couples constrained histories with STEP, views, native files, and text; SldprtNet derives aligned representations from SolidWorks parts; and CADFS reconstructs executable FeatureScript programs with multimodal annotations~\citep{dong2026histcad,li2026sldprtnet,pyatov2026cadfs}.

\datasetname{} occupies a different point in this design space. Its canonical record is a structured feature tree paired with an executable Python view in an \occ{}-based runtime. Standard geometry and learning observations are derived from the accepted construction and linked through release manifests and hashes. The approach favors an accessible executable interface and auditability across representations, while inheriting the distributional limits of a controlled procedural generator.

\section{Dataset Overview}
\label{sec:overview}

\subsection{Design Goals}
\label{subsec:design_goals}

The release design balances three engineering requirements. First, the canonical representation must preserve operation order, parameters, and dependencies, while remaining convertible to code that can be inspected and replayed. Second, every accepted record must resolve to kernel-valid geometry and a standard exchange file; a plausible sequence without a stable solid is insufficient. Third, the representation must remain regular enough for machine learning. The \flluma{} API therefore uses explicit operation names, typed arguments, stable argument order, and direct binding of intermediate objects.

These choices trade unrestricted modeling freedom for scale and alignment. The feature tree, Python view, and training IR describe the same construction at different levels of detail, while derived observations are generated only after acceptance. The result is easier to audit and tokenize than arbitrary user scripts, but it necessarily reflects the operation vocabulary and procedural conventions of the generator.

\subsection{Sample Structure}
\label{subsec:sample_structure}

Each accepted sample contains a coherent set of source, derived, supervision, and provenance files, as summarized in Table~\ref{tab:sample_structure}. The canonical \path{feature_tree.json} records the ordered modeling operations, semantic roles, parameters, dependencies, and topology references, providing construction-history supervision related to that used in Fusion 360 Gallery, DeepCAD, HistCAD, and SldprtNet~\citep{willis2021fusion360gallery,wu2021deepcad,dong2026histcad,li2026sldprtnet}. The paired Python program is the replayable code view of this record and constructs the CAD model in \flluma{}. The feature tree can likewise be replayed in the \flluma{} runtime and converted back into executable Python code for model reconstruction.

The training IR serializes the same construction history into a compact sequence-learning target. Geometry is exported as STEP, while point clouds and canonical views provide geometric and visual supervision. The visual data consist of eight feature-aware visible-edge renderings under \path{images/canonical/}. These views are generated from fixed orthographic and isometric viewpoints and are intended to preserve construction-relevant visible edges rather than photorealistic appearance. Text files provide deterministic descriptions and LLM-enriched prompts, while metadata and release reports record hashes, quality checks, schema version, and generation provenance. At release level, \datasetname{} is distributed as raw sample folders together with JSON release manifests, split files, schema descriptions, vocabulary files, and Parquet training tables, so users can inspect individual CAD records or load train/validation/test splits directly.

\begin{table}[htbp]
  \centering
  \footnotesize
  \caption{Main per-sample and release-level files in \datasetname{}.}
  \label{tab:sample_structure}
  \begin{tabular}{L{0.32\linewidth}L{0.58\linewidth}}
    \toprule
    \textbf{File or directory}         & \textbf{Purpose}                                                                                          \\
    \midrule
    \path{program.py}                  & Executable Python CAD program used to generate and replay the model.                                      \\
    \path{feature_tree.json}           & Structured construction history with operations, parameters, semantic types, dependencies, and selectors. \\
    \path{training_ir.txt}             & Compact training-oriented IR serialized from the feature tree for sequence-learning targets.              \\
    \path{model.step}                  & Kernel-exported STEP geometry for standard CAD exchange and downstream B-Rep processing.                  \\
    \path{point_cloud.npz}             & Surface point cloud with normals and face identifiers.                                                    \\
    \path{images/canonical/}           & Eight feature-aware visible-edge views from fixed orthographic and isometric directions.                  \\
    \path{text/descriptions.json}      & Template-generated textual descriptions and structured CAD facts.                                         \\
    \path{text/llm_descriptions.jsonl} & Per-sample LLM-enriched captions, technical descriptions, and user prompt variants.                       \\
    \path{metadata.json}               & Provenance, hashes, validation summaries, export status, and sample-level statistics.                     \\
    \path{dataset_manifest.json}       & Release-level manifest with split membership, sample paths, modality inventory, and integrity summaries.  \\
    \path{quality_report.json}         & Release-level quality report covering modality completeness, duplicate checks, and operation coverage.    \\
    \path{scheduler_report.json}       & Scheduler report describing template allocation, complexity levels, and generation statistics.             \\
    \path{splits/}                     & Text files listing the official train, validation, and test sample identifiers.                           \\
    \path{schema/} and \path{vocab/}   & Schema descriptions and token vocabularies for the released training representations.                     \\
    \path{parquet/train.parquet}       & Training split table for direct machine-learning data loading.                                            \\
    \path{parquet/val.parquet}         & Validation split table.                                                                                   \\
    \path{parquet/test.parquet}        & Test split table.                                                                                         \\
    \bottomrule
  \end{tabular}
\end{table}

\subsection{Visual Modality}
\label{subsec:overview_visual_views}

The primary visual modality is the canonical eight-view set: six axis-aligned orthographic views and two isometric views. Rather than imitating product photographs, the renderer preserves visible feature boundaries so that holes, pockets, bosses, patterns, and local construction details remain observable from fixed viewpoints. Pure silhouette projections would conceal many internal or back-facing construction cues.

\begin{figure}[t]
  \centering
  \includegraphics[width=\linewidth]{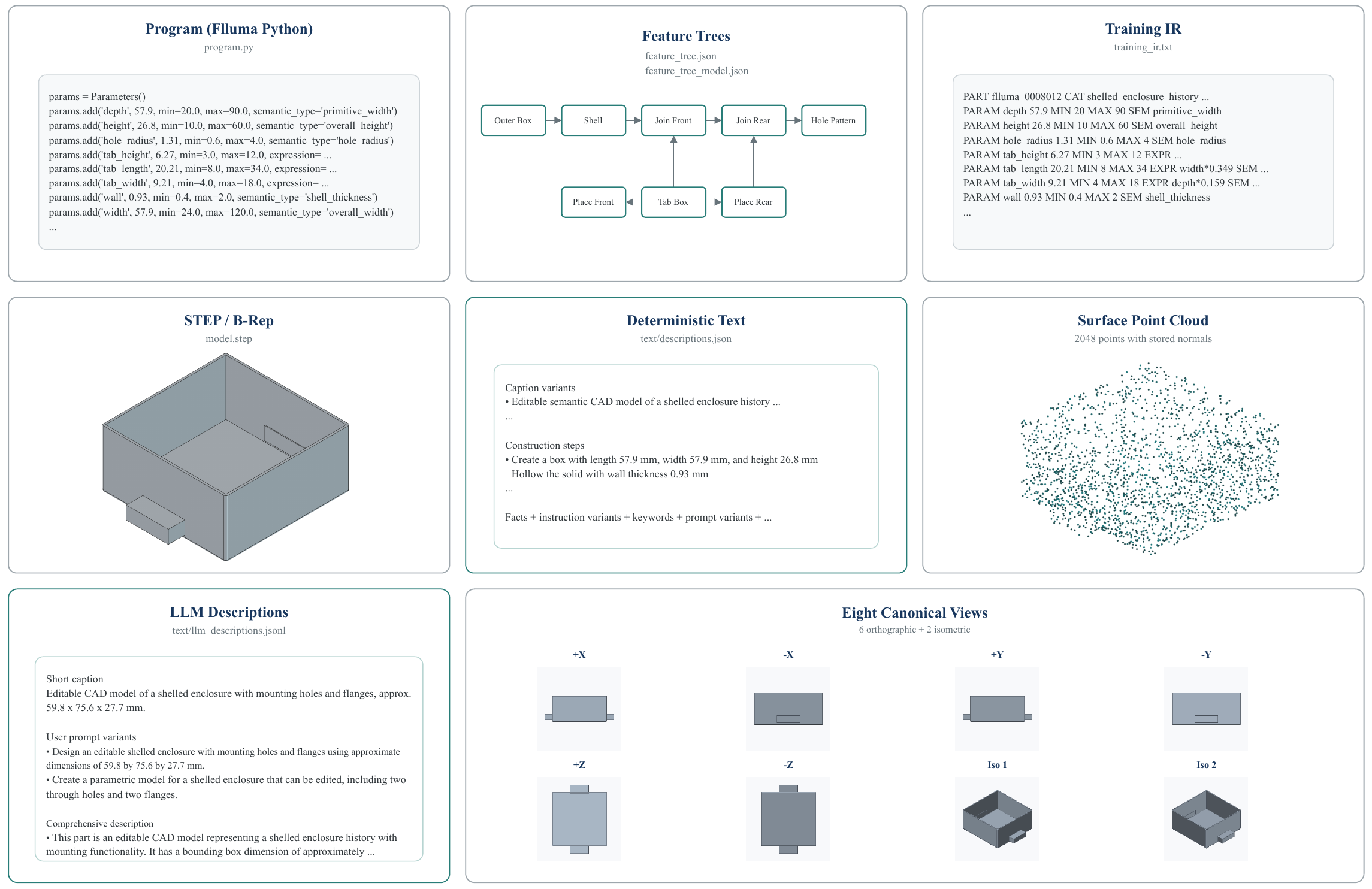}
  \caption{Aligned representations of the released sample \texttt{flluma\_0008012}. The executable Python program, canonical and model-facing feature trees, training IR, STEP B-Rep, 2,048-point surface point cloud with stored normals, deterministic descriptions, LLM-enriched descriptions, and eight canonical views are derived from the same kernel-validated construction record. Textual panels show shortened excerpts; the visual set contains six orthographic and two isometric views.}
  \label{fig:multimodal_alignment}
\end{figure}

Figure~\ref{fig:multimodal_alignment} shows how these views are stored alongside the other representations of one released sample. The panels are not independently collected annotations: the program, construction representations, geometry, point samples, images, and text all refer to the same accepted CAD record. This correspondence allows a model prediction in one representation to be checked against the executable source and kernel-generated geometry in the others.

\subsection{Scale and Complexity Allocation}
\label{subsec:scale_splits_allocation}

The primary release contains \datasetsize{} valid CAD samples. Its scheduler assigns fixed targets to four complexity levels (Table~\ref{tab:level_targets}) so that basic primitives remain represented without allowing them to dominate the corpus, a known risk in command-sequence datasets~\citep{wu2021deepcad,rukhovich2025cadrecode}. The levels are template-level design annotations: each of the 53 template families is assigned one level before generation, and every sample inherits the level of its source template. Feature count, history depth, and topology statistics are measured after generation and are not used as post-hoc level thresholds.

\begin{figure}[t]
  \centering
  \includegraphics[width=\linewidth]{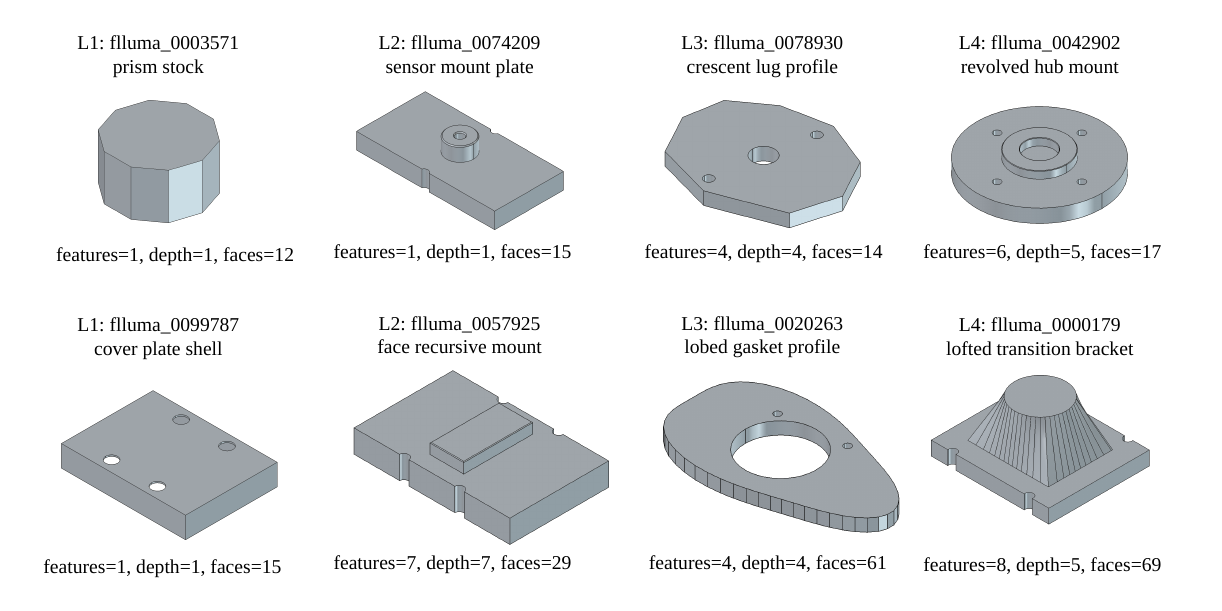}
  \caption{Representative samples from the four complexity levels in \releasename{}. Each level is illustrated with two canonical isometric views together with the sample identifier, template family, feature count, feature-tree depth, and B-Rep face count.}
  \label{fig:complexity_levels}
\end{figure}

L1 covers primitive and low-operation templates used for syntax and geometry grounding. L2 contains basic parametric feature combinations. L3 combines multiple editable features and moderately involved histories. L4 contains templates designed around richer construction sequences, compound feature interactions, and more involved mechanical parts. The labels organize the generator's construction regimes; they are not proposed as a universal geometric complexity scale. Representative samples are shown in Figure~\ref{fig:complexity_levels}.

\begin{table}[htbp]
  \centering
  \small
  \caption{Target and realized complexity allocation for the \releasename{} release.}
  \label{tab:level_targets}
  \begin{tabular}{lrrrr}
    \toprule
    \textbf{Allocation} & \textbf{L1} & \textbf{L2} & \textbf{L3} & \textbf{L4} \\
    \midrule
    Target count   & 3,000       & 17,000      & 55,000      & 25,000      \\
    Target ratio   & 3\%         & 17\%        & 55\%        & 25\%        \\
    Accepted count & 3,056       & 16,966      & 55,219      & 24,759      \\
    \bottomrule
  \end{tabular}
\end{table}

The accepted proportions remain within 0.25 percentage points of their targets.

\subsection{Supported Learning Tasks}
\label{subsec:supported_tasks}

The aligned structure of \datasetname{} enables several learning tasks in editable CAD reconstruction, multimodal CAD generation, and retrieval. These tasks are related to recent work on text-to-CAD generation, multimodal CAD synthesis, CAD program generation, and B-Rep learning~\citep{khan2024text2cad,lv2025cadinstruct,xu2025cadmllm,wang2025cadgpt,rukhovich2025cadrecode,guan2025cadcoder,jayaraman2021uvnet}. Since each sample contains executable code, structured construction history, validated geometry, visual views, point clouds, and text, the dataset can be used from both generative and analytical perspectives.

\begin{itemize}
  \item \textbf{Text-to-CAD and CAD program synthesis:} Models can map natural-language prompts, technical descriptions, or retrieval captions to executable Python CAD programs, feature trees, or training-oriented IRs. The prediction target is editable construction logic rather than only final geometry~\citep{khan2024text2cad,lv2025cadinstruct,guan2025cadcoder}.

  \item \textbf{Image-to-CAD reconstruction:} The canonical visible-edge views provide visual conditioning for reconstructing feature trees, training IRs, or executable CAD programs from fixed CAD views~\citep{alam2025gencad,gao2024diffcad,wang2025cadgpt,xu2025cadmllm}.

  \item \textbf{Point-cloud-to-CAD and editable reverse engineering:} Surface point clouds can be used as input for recovering structured construction histories, typed parameters, or programmatic CAD representations~\citep{rukhovich2025cadrecode,zhang2025ecadnet}.

  \item \textbf{Feature-tree prediction and design auto-completion:} The ordered feature trees and training IRs support prediction of operations, parameters, dependencies, semantic roles, and next-step modeling actions from partial construction histories~\citep{willis2021fusion360gallery,wu2021deepcad,dong2026histcad,li2026sldprtnet}.

  \item \textbf{Cross-modal CAD retrieval:} Text descriptions, rendered views, point clouds, STEP geometry, feature trees, and programs can be embedded into a shared retrieval space, enabling retrieval of editable CAD models from language, images, or geometry~\citep{jayaraman2021uvnet,xu2025cadmllm,wang2025cadfusion}.

  \item \textbf{B-Rep feature extraction and manufacturing-oriented analysis:} Since each sample includes STEP geometry aligned with feature trees and executable programs, the dataset can support research on extracting operation-level or machining-relevant structure from B-Rep geometry, including holes, pockets, bosses, chamfers, fillets, and repeated feature patterns~\citep{jayaraman2021uvnet,colligan2022hierarchicalcadnet,zhang2024brepmfr}.
\end{itemize}

Beyond benchmark construction, these tasks correspond to practical CAD workflows. In editable reverse engineering, a scan-derived point cloud can be mapped back to a feature history that a designer can modify rather than to a static mesh. In manufacturing-oriented analysis, the aligned feature tree and STEP B-Rep can help associate geometric features such as holes, pockets, bosses, chamfers, and fillets with downstream process-planning decisions and operation sequencing.

\section{The \datasetname{} Pipeline}
\label{sec:pipeline}

\begin{figure}[!htb]
  \centering
  \includegraphics[width=\linewidth,height=0.45\textheight,keepaspectratio]{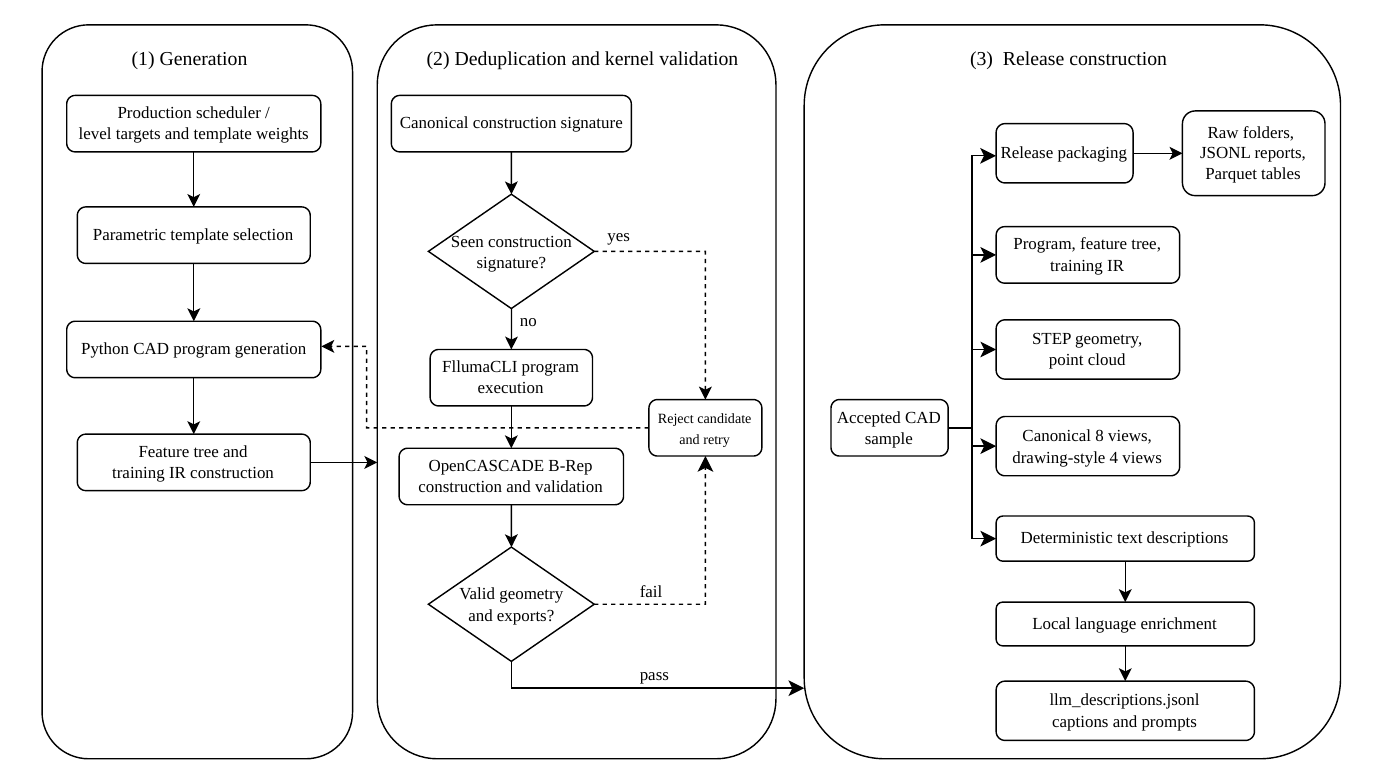}
  \caption{Overview of the \datasetname{} data generation and release pipeline. Parametric CAD programs are generated from scheduled templates, deduplicated using canonical construction signatures, executed and validated in the \flluma{} kernel environment, and then packaged with geometry, visual, textual, and tabular release assets.}
  \label{fig:pipeline}
\end{figure}

Figure~\ref{fig:pipeline} summarizes the generation and release pipeline. The pipeline starts from a scheduled template distribution, generates executable CAD programs and feature trees, filters candidates using design signatures and kernel validation, and then packages accepted samples with geometry, visual views, text, metadata, release reports, split files, and Parquet training tables.

\subsection{\flluma{} Execution Environment}
\label{subsec:flluma_environment}

\flluma{} is a parametric CAD application and execution environment developed to construct, execute, validate, and inspect the code-native samples within \datasetname{}.\footnote{The \flluma{} software ecosystem, including the interactive GUI, headless CLI, and Python CAD API wrappers, is publicly available free of charge for academic research, dataset validation, and program execution at \url{https://www.flluma.com}.} It has a Qt/C++ application layer, a Python-facing CAD API, and an \occ{}-based B-Rep geometry backend. The same system is used in two modes. The graphical interface is used for visual inspection, debugging, and manual checks during development, while \texttt{FllumaCLI} is used for headless batch generation.

In the automated pipeline, \texttt{FllumaCLI} executes each generated Python program, resolves the geometry through the CAD kernel, validates the resulting solid, and produces the release assets. Because the executable program is retained, a sample can be replayed and checked against its exported geometry and construction record rather than inspected only as a STEP file.

The kernel-level validation chain is deliberately explicit. Modeling operations are resolved to \occ{} B-Reps. Boolean joins, cuts, and intersections use OCC Boolean operators from the \texttt{BRepAlgoAPI} family. The resulting shape is checked with the OCC B-Rep check analyzer, including face-, edge-, and vertex-level validity; volume and surface-area facts are computed with \texttt{BRepGProp}; and STEP export/reload checks use \texttt{STEPControl} writer and reader APIs. This makes the validation target a kernel-accepted B-Rep solid, not only a syntactically valid Python script.

Unlike a general-purpose scripting interface, the \flluma{} runtime is used here as part of a fixed release pipeline. CadQuery and FreeCAD accept broader user programs; the present pipeline favors a regular operation schema, explicit dependencies, and deterministic export so that one accepted construction can be traced across its released representations.

\subsection{Program and Feature-Tree Generation}
\label{subsec:program_feature_generation}

Each sample is generated by a Python CAD program executed in the \flluma{} runtime. Treating CAD as executable code follows the direction of CAD-Recode, CAD-Coder, FutureCAD, and CADFS, where program synthesis is used to make generated designs inspectable and editable~\citep{rukhovich2025cadrecode,guan2025cadcoder,li2026futurecad,pyatov2026cadfs}. The programs call typed CAD operations for plate creation, sketch construction, extrusion, cylinders, boxes, Boolean joins and cuts, hole patterns, rectangular pockets, fillets, chamfers, linear and circular patterns, lofts, sweeps, shells, and face-based features.

The Python-facing \flluma{} API is designed as a regularized CAD programming interface for learning. Its commands use explicit operation names, typed parameters, stable argument order, and direct variable binding for intermediate solids, sketches, and feature operations. In practice, this keeps the program text close to the corresponding construction history: a line of code usually introduces one modeling step, one parameterized operation, or one explicit dependency. The goal is not to replace general CAD scripting tools, but to provide a dataset representation whose source code is regular enough for machine learning while remaining executable and editable.

For each accepted model, the pipeline stores both the executable program and a structured feature tree. The program is the human-readable construction source, while the feature tree records ordered operations, parameters, semantic roles, and dependencies in a machine-readable form. These two views are aligned during generation: the feature tree can be replayed in the \flluma{} runtime, and it can also be converted back into executable Python code for model reconstruction. A compact training IR is then derived from the feature tree for sequence-learning targets.

\subsection{Template Scheduling and Complexity Control}
\label{subsec:template_scheduling}

The generator uses a controlled library of parametric templates. Some templates are simple primitives or low-operation examples, while others contain longer editable histories with dependent operations. Template- and grammar-based generation is common in synthetic CAD datasets, but narrow operation vocabularies and conservative parameter ranges can make the resulting data visually repetitive~\citep{wu2021deepcad,rukhovich2025cadrecode,ataei2026zerotocad}. We therefore use a mixture of simple and history-rich templates rather than relying on one construction pattern.

The current library includes primitive stocks, sketch-profile plates, circular flange patterns, motor brackets with cutouts, shelled enclosures, intersected mounting blocks, swept pipe brackets, revolved hub components, lofted transition brackets, face-feature service covers, ribbed heat-sink mounts, and irregular sketch-profile parts. Across these templates, the generator varies parameter values, feature counts, operation order, and semantic roles. The scheduler assigns candidates according to the L1--L4 allocation described in Section~\ref{subsec:scale_splits_allocation}, so that elementary examples are present but most of the release is concentrated in moderate and rich construction histories.

\subsection{Deduplication and Kernel-in-the-Loop Validation}
\label{subsec:dedup_validation}

Generated programs are not accepted only because they are syntactically valid. Before kernel execution, candidates are checked against canonical construction signatures to reduce exact repetition in construction history. Candidates with previously seen design signatures are rejected and regenerated with different seeds or parameters.

Candidates that pass the signature check are executed in \flluma{} and evaluated by the \occ{}-based geometry backend. A candidate is accepted only when the resulting model passes solid validation and the required exports can be produced. This matters because executable CAD programs can fail geometrically even when the code itself is valid, for example through invalid Boolean results, degenerate faces, non-manifold geometry, or unstable topology~\citep{ataei2026zerotocad,guan2025cadcoder,li2026futurecad}. For accepted samples, the pipeline records validation status, export status, source hashes, geometry hashes, and release-level quality summaries. Failed candidates are discarded or retried, and long generation runs use an atomic and resumable protocol so that previously accepted samples can be retained after interruption.

The final manifest records 100,000 evaluated release slots, 77 retries caused by duplicate construction signatures, and no unrecovered generation failures. Transient kernel failures were handled during generation, but their reason categories were not retained as release-level counters; the reported statistics therefore describe accepted outputs and recovered duplicate retries rather than a retrospective failure taxonomy.

\subsection{Multimodal Export}
\label{subsec:multimodal_export}

Accepted samples are exported into several aligned modalities. Geometry is serialized as STEP for standard CAD exchange, and point clouds are sampled from the validated geometry. The visual export produces the canonical eight-view visible-edge set, which exposes local construction-relevant edges from fixed orthographic and isometric viewpoints.

Text supervision is produced in two stages. First, deterministic descriptions are generated from validated sample metadata and written to \path{text/descriptions.json}. These descriptions include the part category, dimensions, operation counts, construction steps, and B-Rep facts. Second, a local LLM enrichment stage rewrites these structured descriptions into more natural Text-to-CAD language. The enrichment stage uses a local Qwen2.5-7B model~\citep{yang2025qwen25}, conditioned on \path{descriptions.json} and the paired \path{training_ir.txt}. The prompt asks for CAD-oriented language while preserving numeric dimensions, operation counts, hole and pattern counts, and bounding-box facts. The resulting \path{text/llm_descriptions.jsonl} file stores a compact retrieval caption, a detailed technical description, and three natural user prompt variants. The LLM stage does not choose CAD parameters, change geometry, repair failed programs, or participate in validation.

\subsection{Release Packaging and Quality Reporting}
\label{subsec:release_packaging_quality}

After sample generation and validation, the release pipeline writes both raw sample folders and machine-learning tables. Raw folders contain the executable program, feature tree, training IR, STEP file, point cloud, visual views, text files, and metadata. Release-level JSON reports index these folders and record split membership, hashes, template labels, operation tokens, modality completeness, and quality information. The same accepted samples are exported into Parquet train, validation, and test tables for direct machine-learning use.

This layout supports two common workflows. A CAD researcher can open a single raw folder and inspect the program, feature tree, geometry, images, and metadata together. A machine-learning user can load the Parquet splits directly without traversing nested sample folders. Both views are derived from the same accepted samples, so a training row can be traced back to its raw CAD record.

The final release report records duplicate sample identifiers, duplicate program hashes, duplicate feature-tree hashes, duplicate training-IR hashes, duplicate solid-signature hashes, missing modalities, failed exports, and warning types. These checks separate failure modes that are easy to confuse in synthetic CAD generation: code that does not execute, code that executes but produces invalid geometry, valid geometry that cannot be exported reliably, derived views that are missing or inconsistent, and accepted samples that duplicate an existing construction record.

\section{Dataset Statistics}
\label{sec:statistics}

This section reports the release-level statistics of \releasename{}. The statistics are computed from the completed 100K release using internal release-analysis reports derived from the raw sample folders, Parquet splits, \path{dataset_manifest.json}, \path{quality_report.json}, and \path{scheduler_report.json}. Following common practice in CAD dataset papers, we report sample count, split integrity, operation coverage, construction-history complexity, B-Rep topology, modality completeness, and duplicate statistics~\citep{koch2019abc,wu2021deepcad,lv2025cadinstruct,dong2026histcad,li2026sldprtnet}.

\begin{figure}[t]
  \centering
  \includegraphics[width=\linewidth]{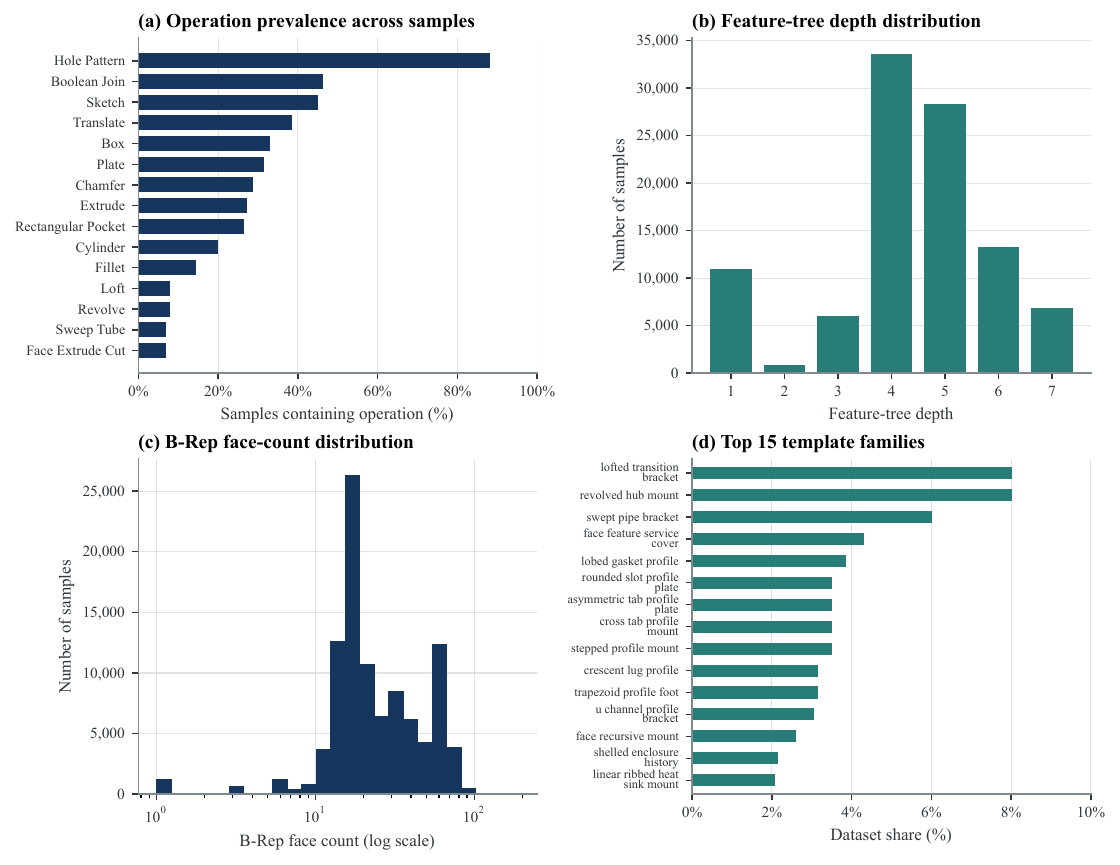}
  \caption{Release-level distributions for \releasename{}. (a) Percentage of samples containing each of the 15 most prevalent operation tokens. (b) Distribution of feature-tree depth. (c) Distribution of B-Rep face count using logarithmic bins and a logarithmic horizontal axis. (d) Dataset share of the 15 most frequent template families.}
  \label{fig:dataset_statistics}
\end{figure}

Figure~\ref{fig:dataset_statistics} provides a distributional overview before the aggregate tables. It reports operation prevalence at the sample level rather than raw token frequency, together with feature-tree depth, B-Rep face-count, and template-family distributions. These views expose both the central construction regime and the long-tailed portions of the release that are less apparent from means and medians alone.

\begin{table}[t]
  \centering
  \small
  \caption{Release-level statistics for \releasename{}.}
  \label{tab:release_statistics}
  \begin{tabular}{ll}
    \toprule
    \textbf{Statistic}                  & \textbf{Value}                      \\
    \midrule
    Total samples scanned               & 100,000                             \\
    Complete samples                    & 100,000 (100.00\%)                  \\
    Train/validation/test split         & 80,000 / 10,000 / 10,000            \\
    Complexity allocation (L1/L2/L3/L4) & 3,056 / 16,966 / 55,219 / 24,759    \\
    Template/category count             & 53                                  \\
    Operation-token vocabulary size     & 44                                  \\
    Feature-tree operation labels       & 60                                  \\
    Mean feature count                  & 5.458                               \\
    Mean history depth                  & 4.348                               \\
    Mean dependency-edge count          & 4.678                               \\
    Mean parameter count                & 6.983                               \\
    STEP files present                  & 100,000 (100.00\%)                  \\
    Point-cloud files present           & 100,000 (100.00\%)                  \\
    Canonical rendered views            & 8 per sample, 100,000 complete sets \\
    Deterministic text descriptions     & 100,000 (100.00\%)                  \\
    Exact duplicate program hashes      & 0                                   \\
    Exact duplicate feature-tree hashes & 0                                   \\
    Exact duplicate training-IR hashes  & 0                                   \\
    Exact duplicate solid signatures    & 0                                   \\
    \bottomrule
  \end{tabular}
\end{table}

\subsection{Complexity Allocation}
\label{subsec:complexity_allocation_statistics}

The production scheduler targets a controlled distribution over four complexity levels. The realized allocation closely matches the intended 3\%/17\%/55\%/25\% ratio, with small deviations caused by validation, retry, and template availability during generation. Table~\ref{tab:realized_level_statistics} reports the measured level counts and representative complexity statistics. L1 contains simple primitives or low-operation examples, while L3 and L4 contain the majority of the release and provide moderate-to-rich construction histories.

\begin{table}[t]
  \centering
  \small
  \caption{Realized complexity allocation and per-level statistics.}
  \label{tab:realized_level_statistics}
  \begin{tabular}{lrrrrr}
    \toprule
    \textbf{Level} & \textbf{Count} & \textbf{Share} & \textbf{Mean features} & \textbf{Mean depth} & \textbf{Mean faces} \\
    \midrule
    L1             & 3,056          & 3.1\%          & 1.000                  & 1.000               & 5.616               \\
    L2             & 16,966         & 17.0\%         & 4.030                  & 4.030               & 22.993              \\
    L3             & 55,219         & 55.2\%         & 5.623                  & 4.412               & 27.392              \\
    L4             & 24,759         & 24.8\%         & 6.620                  & 4.839               & 38.737              \\
    \bottomrule
  \end{tabular}
\end{table}

\subsection{Feature-History and Geometry Complexity}
\label{subsec:feature_geometry_complexity}

In addition to aggregate sample count, \datasetname{} is characterized by construction-history and B-Rep complexity. These statistics are important because a large CAD dataset can still be procedurally narrow if most examples share similar operation counts, dependency structures, or topological patterns. Prior CAD datasets and B-Rep learning papers report command-sequence lengths, topology statistics, sketch constraints, or feature histories to characterize structural difficulty~\citep{koch2019abc,seff2020sketchgraphs,wu2021deepcad,jayaraman2021uvnet,dong2026histcad}.

\begin{table}[t]
  \centering
  \caption{Construction-history and geometry complexity statistics for \releasename{}.}
  \label{tab:complexity_statistics}
  \begin{tabular}{lrrrr}
    \toprule
    \textbf{Metric}           & \textbf{Mean} & \textbf{Median} & \textbf{Min} & \textbf{Max} \\
    \midrule
    Program length (lines)    & 17.442        & 17.000          & 7            & 26           \\
    Feature-tree nodes        & 8.245         & 7.000           & 2            & 16           \\
    Feature count             & 5.458         & 6.000           & 1            & 11           \\
    History depth             & 4.348         & 4.000           & 1            & 7            \\
    Dependency edges          & 4.678         & 5.000           & 0            & 10           \\
    Parameters                & 6.983         & 7.000           & 1            & 13           \\
    Training-IR tokens        & 179.878       & 167.000         & 28           & 307          \\
    B-Rep faces               & 28.789        & 20.000          & 1            & 191          \\
    B-Rep edges               & 72.144        & 48.000          & 2            & 415          \\
    B-Rep vertices            & 46.567        & 32.000          & 1            & 227          \\
    Bounding-box aspect ratio & 6.032         & 4.487           & 1.000        & 28.394       \\
    \bottomrule
  \end{tabular}
\end{table}

The topology statistics show that the release is not limited to single primitive solids. The mean B-Rep topology contains 28.789 faces and 72.144 edges, while the upper tail reaches 191 faces and 415 edges. The median feature count is 6 and the median history depth is 4, indicating that the central portion of the dataset contains editable multi-step construction histories rather than only isolated final shapes.

Each accepted sample is exported as STEP using the OpenCASCADE STEP writer with the AP214 automotive-design schema. The aligned point cloud contains 2,048 float32 surface samples with float32 normals. Points are generated by the \flluma{} surface-UV sampler: sample counts are allocated across CAD faces in proportion to face area, with small-face preservation enabled and at least one point assigned to each face when feasible. Within each face, candidate points are drawn in the trimmed UV domain, filtered by the face classifier, and accepted according to the local surface Jacobian so that sampling better follows surface area than raw UV coordinates. Normals are computed from the cross product of the first derivatives of the underlying \occ{} surface and adjusted for face orientation. The point-cloud export records this policy as \path{sampling=surface_uv} in sample metadata and release schema files.

\subsection{Operation and Template Diversity}
\label{subsec:operation_feature_diversity}

The release contains 53 template or category families and 44 operation tokens. The ten most frequent template families cover 47.4\% of the dataset, and the largest family, \texttt{lofted\_transition\_bracket}, accounts for 8.0\%. The template distribution has a normalized entropy of 0.915 and a Gini coefficient of 0.446. Normalized entropy equals 1 for a uniform distribution over the 53 families; the measured value shows that samples are spread across the catalog, although several recurring mechanical templates remain visibly concentrated.

Two related vocabularies are reported separately. In the 44-token model-facing operation vocabulary, \texttt{OP\_HOLE\_PATTERN} appears in 88.1\% of samples and accounts for 17.8\% of operation-token instances. The ten most frequent tokens cover 82.2\% of instances; loft, revolve, sweep, shell, circular pattern, threaded hole, and profile cut occupy the long tail. The feature-tree representation uses a finer 60-label vocabulary containing semantic and composite construction nodes. Under this counting scheme, \texttt{HolePattern} is the most frequent label at 14.2\%, and the label distribution has a normalized entropy of 0.747 and a Gini coefficient of 0.734. The two percentages use different denominators. Their imbalance reflects the prevalence of mounting-hole and pattern operations in mechanical parts, so operation-level models should report macro-averaged or per-operation results in addition to micro-averaged accuracy.

\begin{table}[t]
  \centering
  \caption{Diversity statistics for \releasename{}.}
  \label{tab:diversity_statistics}
  \begin{tabular}{ll}
    \toprule
    \textbf{Statistic}                    & \textbf{Value}                       \\
    \midrule
    Template/category count               & 53                                   \\
    Operation-token vocabulary size       & 44                                   \\
    Feature-tree operation-label count    & 60                                   \\
    Most frequent template/category       & \texttt{lofted\_transition\_bracket} \\
    Most frequent template/category ratio & 8.0\%                                \\
    Top-10 template/category coverage     & 47.4\%                               \\
    Template normalized entropy           & 0.915                                \\
    Template Gini coefficient             & 0.446                                \\
    Most frequent feature-tree label      & \texttt{HolePattern}                 \\
    Most frequent feature-tree label ratio & 14.2\%                              \\
    Feature-tree label normalized entropy & 0.747                                \\
    Feature-tree label Gini coefficient   & 0.734                                \\
    \bottomrule
  \end{tabular}
\end{table}

\subsection{Modal Alignment and Integrity}
\label{subsec:modal_alignment_integrity}

For each sample, hashes and schema checks verify consistency among the executable program, feature tree, training IR, metadata, STEP geometry, point cloud, deterministic text, and canonical rendered views. This integrity reporting is especially important for multimodal CAD releases, where missing views, inconsistent programs, or invalid B-Reps can affect multiple downstream tasks~\citep{koch2019abc,lv2025cadinstruct,xu2025cadmllm,dong2026histcad,li2026sldprtnet}.

\begin{table}[t]
  \centering
  \caption{Release integrity metrics for \releasename{}.}
  \label{tab:integrity_metrics}
  \begin{tabular}{ll}
    \toprule
    \textbf{Metric}                      & \textbf{Value}     \\
    \midrule
    Complete samples                     & 100,000 (100.00\%) \\
    Incomplete or unstable samples       & 0                  \\
    STEP completeness                    & 100,000 (100.00\%) \\
    Point-cloud completeness             & 100,000 (100.00\%) \\
    Canonical 8-view completeness        & 100,000 (100.00\%) \\
    Deterministic-text completeness      & 100,000 (100.00\%) \\
    Duplicate program hashes             & 0                  \\
    Duplicate feature-tree hashes        & 0                  \\
    Duplicate training-IR hashes         & 0                  \\
    Duplicate solid-signature hashes     & 0                  \\
    Duplicate-signature retries          & 77                 \\
    Unrecovered generation failures      & 0                  \\
    Missing required files by type       & None               \\
    Unstable or unreadable files by type & None               \\
    \bottomrule
  \end{tabular}
\end{table}

No exact duplicates were found under program, feature-tree, training-IR, metadata, geometry, solid-signature, or shape-signature hashes. This does not mean that no two samples are visually similar: samples from the same template family may share a global layout while differing in dimensions, local feature counts, operation parameters, topology, or construction history. For this reason, the release reports exact duplicate hashes as hard integrity checks, while template concentration and operation distributions are used to characterize dataset diversity.

\subsection{Text Supervision}
\label{subsec:text_quality_statistics}

The present release scan includes deterministic descriptions as the main auditable text modality. These descriptions are complete for all 100,000 samples and are generated directly from validated metadata, construction histories, and B-Rep facts. Their mean length is approximately 9,817 characters, with a median of 10,011 characters.

The final release also includes LLM-enriched records for all samples in \path{text/llm_descriptions.jsonl}. We keep deterministic descriptions and LLM-enriched prompts conceptually separate. The deterministic text is used for reproducible dataset auditing, while the LLM-enriched captions, technical descriptions, and user-prompt variants provide an additional instruction-style supervision layer. Kernel-derived facts are therefore not mixed with generated paraphrases in the core integrity statistics.

\section{Benchmark Protocol and Baseline Study}
\label{sec:benchmarks}

We define benchmark tasks over the released representations and report a text-to-program experiment on the official split. The experiment tests the complete path from a language description to executable code, kernel validation, STEP export, and geometric comparison.

\subsection{Benchmark Tasks}
\label{subsec:benchmark_tasks}

\begin{table}[t]
  \centering
  \caption{Proposed benchmark tasks for \datasetname{}.}
  \label{tab:benchmarks}
  \begin{tabular}{L{0.23\linewidth}L{0.36\linewidth}L{0.29\linewidth}}
    \toprule
    \textbf{Task}               & \textbf{Input / output}                                  & \textbf{Metrics}                                        \\
    \midrule
    Template classification     & Image, point cloud, or text to template family           & Accuracy, macro-F1                                      \\
    Complexity-level prediction & Image, point cloud, or training IR to L1--L4 level       & Accuracy, macro-F1, confusion matrix                    \\
    Next-operation prediction   & Prefix of training IR to next operation token            & Top-1, top-5 accuracy                                   \\
    Feature-tree reconstruction & Image, point cloud, or text to feature-tree sequence     & Operation accuracy, parameter error, tree-edit distance \\
    CAD retrieval               & Text, image, or point cloud query to matching CAD sample & Recall@K, median rank                                   \\
    Program validity            & Generated program or IR to executable CAD model          & Syntax rate, kernel-valid rate, export-valid rate       \\
    \bottomrule
  \end{tabular}
\end{table}

The task definitions admit standard sequence, image, point-cloud, and text encoders, allowing later work to compare model classes without changing the data contract. Table~\ref{tab:benchmarks} specifies the input, target, and principal metric for each task.

\subsection{Text-to-Program Baseline}
\label{subsec:text_to_program_baseline}

To verify the code-native use of the dataset, we train a text-to-program baseline in which a model maps a natural-language CAD description to an executable \path{program.py}. In addition to recovering the modeling operations and parameters, the output must be syntactically valid Python, use the \flluma{} CAD API correctly, and execute to a valid solid model. The experiment therefore evaluates the main representation choice of \datasetname{}: editable CAD supervision is provided as executable source code rather than only as final geometry or a latent command sequence.

We fine-tune Qwen2.5-Coder-1.5B-Instruct~\citep{yang2025qwen25} with LoRA on the official 80,000-sample training split. For each sample, a deterministic description from \path{text/descriptions.json} or the LLM-generated comprehensive description in \path{text/llm_descriptions.jsonl} is selected with equal probability using a fixed seed; the target is the paired \path{program.py}. Prompt tokens are masked from the causal language-model loss. The model is trained for one epoch with a maximum context length of 4096 tokens, learning rate $2\times10^{-4}$, cosine scheduling, warmup ratio 0.03, LoRA rank 16, LoRA alpha 32, dropout 0.05, and an effective batch size of 8. Training uses FP16 with TF32 enabled and takes 27,318 seconds on one 16-GB NVIDIA RTX 5060 Ti. Greedy decoding is used at test time with at most 2,048 generated tokens. The held-out test set contains 10,000 samples and is never used for training. Release integrity checks report zero exact duplicate program, feature-tree, training-IR, and solid-signature hashes across the splits. The official split is sample-disjoint but not template-disjoint: template families occur in all splits, so this baseline measures interpolation within the released generator distribution rather than generalization to unseen template families. For reproducibility, the GitHub repository provides the data-preparation, LoRA training, generation, Python-level evaluation, and \flluma{} execution-evaluation scripts, together with the run-command configuration and lightweight result summaries including \path{training_report.json}, \path{generation_report.json}, \path{evaluation_python_summary.json}, and \path{evaluation_flluma_summary.json}.

\subsection{Evaluation Metrics}
\label{subsec:baseline_metrics}

The text-to-program task is evaluated at code, execution, and geometry levels. Code-level checks test Python syntax and the required \path{part} object. Execution metrics record \flluma{} loading, model construction, solid validation, and STEP export. For geometric comparison, the reference bounding box center is subtracted from both 2,048-point clouds, and both are divided by half of the reference box's largest side length. The normalized reference therefore spans two units along its longest axis. Chamfer Distance is the mean of the two directed mean squared nearest-neighbor distances. Source-code similarity uses Python's \texttt{difflib.SequenceMatcher} ratio on the raw extracted program strings. Operation precision, recall, and F1 use multiset overlap between extracted \path{part.*} API call names; the operation LCS ratio separately measures their order relative to the reference sequence. These operation metrics do not evaluate argument values, and the precision/recall/F1 scores do not penalize call reordering, so they are interpreted alongside source similarity, execution, and Chamfer results.

The validation loss in Table~\ref{tab:text_to_program_baseline} is measured on a fixed 1,000-sample subset after the epoch. Its value is lower than the epoch-averaged training loss because training uses LoRA dropout and the reported training loss also includes earlier, less-fitted optimization steps. The validity, similarity, and geometry metrics are computed on the full 10,000-sample test split.

\begin{table}[t]
  \centering
  \caption{Text-to-program LoRA baseline on the official 10,000-sample test split.}
  \label{tab:text_to_program_baseline}
  \begin{tabular}{ll}
    \toprule
    \textbf{Metric} & \textbf{Value} \\
    \midrule
    Training samples & 80,000 \\
    Test samples & 10,000 \\
    Train loss / validation-subset loss & 0.0878 / 0.0766 \\
    Python syntax-valid rate & 99.98\% \\
    Required \path{part} definition rate & 99.98\% \\
    \flluma{} load rate & 99.97\% \\
    \flluma{} build rate & 99.97\% \\
    Kernel solid-valid rate & 99.14\% \\
    STEP-export-valid rate & 99.14\% \\
    Chamfer-evaluated samples & 9,909 \\
    Mean / median Chamfer Distance & 0.002124 / 0.001491 \\
    P90 / P95 Chamfer Distance & 0.003502 / 0.004854 \\
    Maximum Chamfer Distance & 0.180119 \\
    Mean source-code similarity & 87.34\% \\
    Operation precision / recall / F1 & 99.97\% / 99.95\% / 99.96\% \\
    Operation LCS ratio & 99.94\% \\
    \bottomrule
  \end{tabular}
\end{table}

Table~\ref{tab:text_to_program_baseline} shows that the baseline produces executable programs for most held-out descriptions. Out of 10,000 test cases, 9,998 are syntactically valid Python programs and 9,997 can be loaded and built by the \flluma{} runtime. Of the built programs, 83 fail solid validation; the remaining 9,914 are solid-valid and all are exported successfully as STEP. Of these predictions, 9,909 are also converted to generated surface point clouds for geometric comparison. The mean and median normalized Chamfer Distances are 0.002124 and 0.001491, respectively, with a P95 value of 0.004854. The maximum value is 0.180119, showing that rare geometric outliers remain despite the low central values. The five-case gap between STEP validity and Chamfer coverage comes from models rejected by the stricter point-cloud export path. The two syntax-invalid programs and one additional load failure account for the earlier execution failures. The high operation F1 is expected under this benchmark design because the multiset metric compares names from a regular CAD API and does not score argument values or call order; order is measured separately by the LCS ratio, while parameter and geometry errors are reflected by source-code similarity, \flluma{} execution, solid validation, STEP export, and Chamfer Distance. These results establish a closed evaluation path from text to executable CAD code and kernel-checked geometry. They should be interpreted as reconstruction within a controlled API and shared template distribution, not as evidence of open-domain CAD generation or unseen-template generalization.

\begin{table}[t]
  \centering
  \caption{Baseline results grouped by the template-level complexity annotation. Chamfer statistics use the successfully sampled predictions in the \textbf{CD $n$} column.}
  \label{tab:baseline_by_level}
  \resizebox{\linewidth}{!}{%
    \begin{tabular}{lrrrrrrrr}
      \toprule
      \textbf{Level} & \textbf{Test $n$} & \textbf{Build} & \textbf{STEP valid} & \textbf{Code sim.} & \textbf{Op. F1} & \textbf{CD $n$} & \textbf{Mean CD} & \textbf{Median CD} \\
      \midrule
      L1 & 321   & 100.00\% & 100.00\% & 83.13\% & 100.00\% & 320   & 0.002906 & 0.001668 \\
      L2 & 1,682 & 99.94\%  & 99.05\%  & 84.12\% & 99.93\%  & 1,665 & 0.001570 & 0.001235 \\
      L3 & 5,513 & 99.98\%  & 99.55\%  & 86.83\% & 99.95\%  & 5,485 & 0.002061 & 0.001292 \\
      L4 & 2,484 & 99.96\%  & 98.19\%  & 91.18\% & 100.00\% & 2,439 & 0.002541 & 0.002012 \\
      \bottomrule
    \end{tabular}%
  }
\end{table}

\FloatBarrier

Table~\ref{tab:baseline_by_level} separates the test results using the level inherited from each sample's template. L4 has the lowest STEP-valid rate and the highest median Chamfer Distance, although its source-code similarity is also the highest. The latter reflects regularity within those template families rather than lower geometric difficulty. The non-monotonic Chamfer values likewise show that the four levels describe construction regimes, not a scalar ranking of reconstruction error.

We also evaluate description-source robustness without retraining. Using the same mixed-trained LoRA adapter but replacing the test inputs with only LLM-generated comprehensive descriptions yields 99.98\% Python syntax validity, 99.96\% \flluma{} build success, 99.18\% STEP-export-valid rate, 99.92\% operation F1, and 87.96\% mean source-code similarity on the same 10,000-sample test split. The small differences from Table~\ref{tab:text_to_program_baseline} indicate that the adapter is not tied to the deterministic description format. This is a test-time description-source comparison, not a separately trained LLM-only baseline.

\section{Comparison with Existing Datasets}

Table~\ref{tab:comparison} compares \datasetname{} with representative geometry repositories, construction-history datasets, multimodal CAD collections, and executable-code datasets~\citep{wu2015modelnet,chang2015shapenet,koch2019abc,willis2021fusion360gallery,wu2021deepcad,lv2025cadinstruct,xu2025cadmllm,dong2026histcad,li2026sldprtnet,rukhovich2025cadrecode,ataei2026zerotocad,pyatov2026cadfs}. Each scale and capability entry follows the corresponding cited paper. A released CAD file is not treated as evidence of program executability or kernel-level validation, and operation counts are given only when the cited paper explicitly reports them.

\begin{table}[t]
  \centering
  \caption{Comparison with representative CAD datasets using the values reported in the cited papers. ``NR'' denotes a field not explicitly quantified by the cited paper; ``N/A'' denotes a field not applicable to geometry-only collections.}
  \label{tab:comparison}
  \renewcommand{\arraystretch}{0.78}
  \resizebox{\textwidth}{!}{%
    \begin{tabular}{L{0.17\linewidth}L{0.13\linewidth}L{0.25\linewidth}L{0.20\linewidth}L{0.30\linewidth}L{0.29\linewidth}}
      \toprule
      \textbf{Dataset} & \textbf{Scale} & \textbf{Program / history representation} & \textbf{Reported operation scope} & \textbf{Released or aligned modalities} & \textbf{Reported execution / validation} \\
      \midrule
      ModelNet & 151,128 & Final 3D assets & N/A & Meshes / voxels and labels & N/A \\
      ShapeNet & $>3$M indexed & Final 3D assets & N/A & 3D models and semantic annotations & N/A \\
      ABC & 1M & Final B-Rep geometry without construction history & N/A & STEP / Parasolid, meshes, and curve / patch labels & OpenCascade processing and filtering \\
      Fusion 360 (reconstruction) & 8,625 & Human-authored sequences & Sketch and extrude & Sequences and target geometry & Fusion 360 Gym \\
      DeepCAD & 178,238 & Vectorized command sequences & Line, arc, circle, extrude & Sequences; derived geometry & B-Rep / point conversion for evaluation \\
      CAD\-Instruct & NR & Natural-language-guided CAD programs & NR & Natural language and CAD programs & NR \\
      Omni-CAD / CAD-MLLM & 453,220 & DeepCAD-style sequences & Line, arc, circle, extrude & Text, eight views, points, sequences & PythonOCC; topology / enclosure metrics \\
      HistCAD & 170,236 & Constraint-aware executable history & Sketch constraints; extrude, revolve, sweep, fillet, chamfer & Sequences, native CAD, STEP, views, text & Execution and editability evaluation \\
      SldprtNet & 242,606 & SolidWorks history as structured text & 13 command types & SLDPRT, STEP, seven views, text & SolidWorks API round trip \\
      CADFS & 450k & Executable FeatureScript & 15 modeling operations & Programs and multimodal annotations & Reconstructed and executed programs \\
      CAD-Recode & 1M synthetic & Executable Python programs & Sketch and extrude & Python and point-cloud inputs & CadQuery execution \\
      Zero-to-CAD & $\sim$1M synthetic & Executable readable code & Beyond sketch-extrude; count NR & Sequences and CAD models & Agentic execution and validation \\
      \datasetname{} & 100K & Canonical feature tree and executable Python program & 44 operation tokens; 60 feature labels & Python, feature tree, training IR, STEP, points with normals, eight views, and text & \flluma{} replay, \occ{} solid validation, and STEP export \\
      \bottomrule
    \end{tabular}
  }
\end{table}

The comparison separates representation from scale. Although smaller than the largest program collections, \datasetname{} links its feature tree, executable Python, geometry, and learning observations to one accepted record, allowing failures to be traced across code, construction, kernel output, export, and text.

\FloatBarrier

\section{Limitations}

\datasetname{} is generated by a controlled procedural pipeline, so its distribution reflects the choices of template families, parameter ranges, level allocation, and validation constraints. This is a common trade-off in synthetic CAD and code-based CAD datasets, where scale and controllability can introduce generator-specific biases~\citep{rukhovich2025cadrecode,ataei2026zerotocad,pyatov2026cadfs}. Although history-rich templates and irregular sketch profiles improve diversity, the dataset may still underrepresent the full diversity of human CAD workflows, especially long industrial assemblies, surface-heavy consumer products, and manually authored design histories. The present release uses a local LLM only for text enrichment; it is not an autonomous geometric agent and does not perform closed-loop CAD repair, unlike recent agentic or reward-driven CAD generation pipelines~\citep{ataei2026zerotocad,guan2025cadcoder,li2026futurecad}. Text descriptions may still contain awkward phrasing or fact mismatches, so deterministic and LLM-enriched text should be audited separately, following the broader need for text-CAD consistency in language-guided CAD datasets~\citep{lv2025cadinstruct,khan2024text2cad,li2026sldprtnet}. Finally, executable Python programs improve accessibility and reproducibility, but models trained on the dataset may learn the conventions of the \flluma{} API and program schema. Future releases should therefore include broader template families, stronger text-faithfulness checks, and additional benchmarks that measure generalization beyond the generator distribution.

\section{Conclusion}

\datasetname{} addresses a practical gap between final-shape datasets and proprietary native CAD histories by releasing an inspectable construction record with a replayable code view. The controlled generator makes 100K-scale alignment and kernel validation feasible, at the cost of inheriting its template and API distribution. The baseline confirms that the release supports a reproducible path from language to executable, geometrically comparable CAD output. The most important next step is evaluation under parameter edits and template-disjoint splits, where validity alone is no longer sufficient and preservation of construction intent becomes measurable.

\section*{Data and Code Availability}

The project repository is available at \url{https://github.com/Cad-Kernel/FllumaOne-100K}. It provides dataset documentation, download instructions, and the text-to-program reproducibility package used in this paper. The package includes scripts for data preparation, LoRA training, generation, Python evaluation, \flluma{} execution evaluation, and Chamfer geometry evaluation, together with the run configuration, prompt-source comparison, and summary JSON files used for Table~\ref{tab:text_to_program_baseline}. The downloadable \releasename{} archive contains the complete dataset, release manifests, schemas, split definitions, validation reports, vocabulary files, and license information. The \flluma{} CAD program used to execute and test generated programs is available at \url{https://www.flluma.com}.
\bibliographystyle{unsrtnat}
\bibliography{references}

\end{document}